\documentclass{article}

\PassOptionsToPackage{numbers, compress}{natbib}

\usepackage[final]{neurips_2024}

\usepackage[utf8]{inputenc} 
\usepackage[T1]{fontenc}    
\usepackage{hyperref}       
\usepackage{url}            
\usepackage{booktabs}       
\usepackage{amsfonts}       
\usepackage{nicefrac}       
\usepackage{microtype}      
\usepackage{xcolor}         

\usepackage{graphicx}
\usepackage{wrapfig}
\usepackage{bbding}
\usepackage{tabularray}
\usepackage{subcaption}
\usepackage{amsmath, amssymb}
\usepackage{hyperref}
\usepackage{mathtools}
\usepackage{enumitem}
\usepackage{fontawesome5}

\title{Real-time Stereo-based 3D Object Detection for Streaming Perception}

\author{%
	Changcai Li$^{1, 2}$ \quad Zonghua Gu$^{3}$ \quad Gang Chen$^{1, 2,}$\thanks{Corresponding author} \quad \textbf{Libo Huang}$^4$ \\ 
	\textbf{Wei Zhang}$^2$ \quad \textbf{Huihui Zhou}$^2$
	\\
	$^1$Sun Yat-sen University \quad $^2$Pengcheng Laboratory \\
	$^3$Hofstra University \quad $^4$National University of Defense Technology \\
	\texttt{lichc5@mail2.sysu.edu.cn} \quad	\texttt{zonghua.gu@hofstra.edu} \quad 	\texttt{cheng83@mail.sysu.edu.cn}\href{mailto:cheng83@mail.sysu.edu.cn}{\textsuperscript{\faEnvelope}} \\
	\texttt{libohuang@nudt.edu.cn} \quad \texttt{zhangwei1213052@126.com} \quad 	\texttt{zhouhh@pcl.ac.cn} \\
}

\begin{document}
	\maketitle
	\begin{abstract}
	The ability to promptly respond to environmental changes is crucial for the perception system of autonomous driving. 
	Recently, a new task called streaming perception was proposed. 
	It jointly evaluate the latency and accuracy into a single metric for video online perception. 
	In this work, we introduce StreamDSGN, the first real-time stereo-based 3D object detection framework designed for streaming perception. 
	StreamDSGN is an end-to-end framework that directly predicts the 3D properties of objects in the next moment by leveraging historical information, thereby alleviating the accuracy degradation of streaming perception.
	Further, StreamDSGN applies three strategies to enhance the perception accuracy: 
	(1) A feature-flow-based fusion method, which generates a pseudo-next feature at the current moment to address the misalignment issue between feature and ground truth. 
	(2) An extra regression loss for explicit supervision of object motion consistency in consecutive frames. 
	(3) A large kernel backbone with a large receptive field for effectively capturing long-range spatial contextual features caused by changes in object positions. 
	Experiments on the KITTI Tracking dataset show that, 
	compared with the strong baseline, StreamDSGN significantly improves the streaming average precision by up to 4.33\%.
	Our code is available at \href{https://github.com/weiyangdaren/streamDSGN-pytorch}{\url{https://github.com/weiyangdaren/streamDSGN-pytorch}.}
\end{abstract}
	\section{Introduction}
\label{sec:intro}

\begin{figure*}[htbp]

	\centering
	\includegraphics[width=0.8\linewidth]{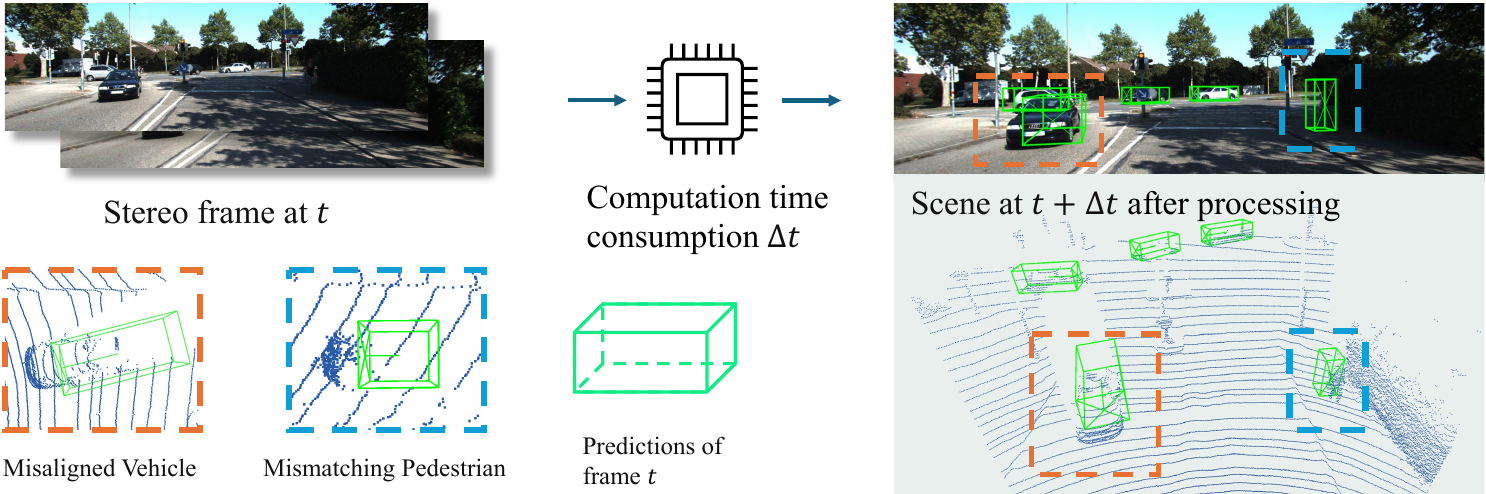}
	\caption{In the context of online streaming perception, the environment changes during inference.}
	\label{fig:offline}

\end{figure*}

Stereo-based 3D object detection~\cite{wang2019pseudo, chen2020dsgn, chen2022dsgn++} presents a notable advantage of low-cost in the deployment compared to LiDAR-based methods.
However, existing stereo-based methods encounter challenges in delivering timely responses due to the intensive computations. Although there have been efforts towards lightweight approaches~\cite{liu2021yolostereo3d, meng2023efficient}, the latency of the inference process is still non-negligible. This is because the latency will lead to the misalignment between the prediction of the latest frame and the real-time changing scene, as shown in Figure~\ref{fig:offline}.

Recently, a new metric~\cite{li2020towards, wang2023we} named streaming accuracy was proposed for real-time online perception. Different from previous offline metrics~\cite{geiger2012we, caesar2020nuscenes, sun2020scalability} which only emphasize the detection accuracy, 
it simultaneously considers both accuracy and latency in the performance evaluation.
With this metric, the model is forced to evaluate the processed world state against the prediction of the received frame. Hence, many high-accuracy but low-efficiency detectors~\cite{lin2017focal, he2017mask} will result in significant performance degradation for real-time perception applications.



\paragraph{Challenges.}

\begin{wrapfigure}{r}{0.5\textwidth}
	
	\centering
	\includegraphics[width=0.5\textwidth]{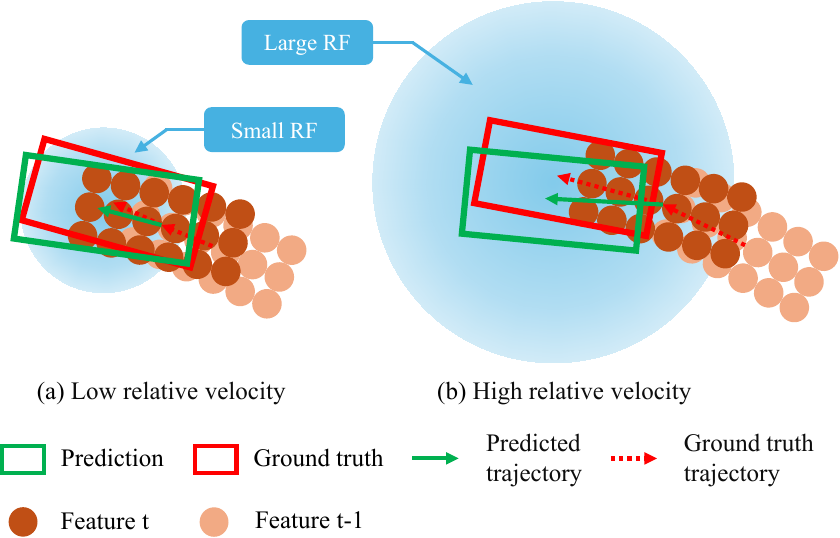}
	\caption{Illustration of the challenges.}
	\label{fig:misalignment}

\end{wrapfigure}


A widely recognized solution for addressing streaming perception is to combine historical features and directly predict future outcomes at the current moment within a real-time framework~\cite{yang2022real, li2023longshortnet, he2023damo}. This is because when the model's inference speed exceeds the input frame rate, the metric of streaming accuracy consistently matches and evaluates the results of the current frame with the ground truth of the next frame. 
However, several challenges arise in the implementation of this solution. As delineated in Figure~\ref{fig:misalignment}, these challenges include the following three specific aspects:

The \textbf{first challenge} is the misalignment between the future supervisory signals and the current features. 
It is observed that for moving objects, their ground truth positions in the next frame at time step $t+1$ (depicted by the \textcolor{red}{red} bounding box) consistently differ from their current positions at time step $t$.  
In such cases, the model needs to learn geometric information about the foreground from distant historical features. Furthermore, this misalignment will exacerbate with the increasing disparity in relative velocities between the ego vehicle and the surrounding vehicles.

The \textbf{second challenge} lies in the implicit supervision when only using the ground truth of a single future frame. It is observed that motion objects with diverse relative velocities exhibit distinct trajectory lengths (depicted by the \textcolor{red}{red} arrow). Previous works~\cite{yang2022real, li2023longshortnet, he2023damo} lack explicit trajectory supervision, requiring models to spontaneously learn various object motion offsets within latent space. Further, due to the complexity of 3D object detection tasks, a single regression supervision method may not be sufficient to ensure accurate detection of objects moving at different speeds.\footnote{In offline perception, this issue does not arise, since the ground truth is synchronized with each input frame.}

The \textbf{third challenge} arises from the effective utilization of context information embedded in the combined features. Due to the generally lower frame rates of 3D datasets compared to 2D ones, for example, the KITTI Tracking dataset~\cite{geiger2012we} has a frame rate of only 10Hz while Argoverse-HD~\cite{li2020towards} has 30Hz. Therefore, in 3D datasets, larger intervals will result in larger spatial distances between the same objects in adjacent frames. At this point, feature extractor with small receptive field (RF) (depicted by the \textcolor{cyan}{cyan} areas) is insufficient to capture information from distant historical features.

\paragraph{Our solution.} 
In this paper, we propose StreamDSGN, a streaming stereo-based 3D detector based on the advanced DSGN++~\cite{chen2022dsgn++} architecture. 
It directly predicts the 3D properties of the \textbf{next} frame by leveraging the fusion of bird's-eye view (BEV) features from both the current and historical frames.
Further, StreamDSGN combines three strategies to tackle the above-mentioned challenges as follows:

\begin{itemize}[leftmargin=2em]
\item[\( \bullet \)] \textit{Feature-Flow Fusion (FFF).} To address the \textbf{first challenge} of misalignment between features and ground truth, we introduce a novel fusion method based on feature flow. It computes the feature flow between the current frame and the previous frame through similarity matching, and subsequently warps the current feature into the BEV coordinates of the next frame according to the flow map. This yields pseudo-next features aligned with the ground truth of the next frame.
\item[\( \bullet \)] \textit{Motion Consistency Loss function (MCL).} To address the \textbf{second challenge} regarding implicit supervision, we establish additional explicit supervision based on motion consistency between adjacent frames. Specifically, we utilize the historical ground truth trajectories to supervise the predicted trajectory, guiding the model to recognize the offset magnitude and direction of object positions in the next frame.
\item[\( \bullet \)] \textit{Large Kernel BEV Backbone (LKBB).} To address the \textbf{third challenge} involving large-span contextual information, we adopt a large receptive field backbone to extract fused features. This structure follows existing research on large kernel convolutions~\cite{guo2023visual, lau2024large}, aiming to enhance the model's capacity to capture long-range dependencies.
	
\end{itemize}
We perform streaming simulations~\cite{li2020towards, wang2023we} and conduct comprehensive experiments on the KITTI Tracking dataset~\cite{geiger2012we}, showing significant improvements in the streaming perception task. To the best of our knowledge, our work represents the \textbf{first} endeavor to explore the implementation of 3D object detection for streaming perception. 

\section{Related Works}
\paragraph{Stereo-based 3D object detection.} 
The methods in this field can be broadly classified into three categories as follows:

1) \textit{2D-detection-based methods.} These methods~\cite{li2019stereo, sun2020disp, xu2020zoomnet, pon2020object, peng2020ida} typically begin by feeding stereo image pairs into a 2D detector or an instance segmentation task to 
generate prior information before performing 3D object detection.
The works in~\cite{liu2021yolostereo3d, peng2022side} implement a single-stage stereo detection framework. Furthermore, \citet{wu2023semi} employ a semi-supervised method to enhance the foundational model by adopting plentiful unannotated images.

2) \textit{Pseudo-LiDAR-based methods.} These methods~\cite{wang2019pseudo, you2020pseudo, qian2020end, garg2020wasserstein} employ the stereo matching task to obtain depth information from the scene and then transform it into a data structure resembling a LiDAR point cloud. 
Further, \citet{li2020confidence} leverage the confidence derived from semantic segmentation to enhance the pseudo-LiDAR. 
The work in~\cite{li2022real, meng2023efficient} adopt a binary neural network (BNN) to quantize the disparity estimator for significantly reducing the computational overhead.

3) \textit{Geometric-volume-based methods.} The methods in~\cite{chen2020dsgn, wang2021plumenet, guo2021liga, liu2023stereodistill, chen2022dsgn++} adopt a representation that encodes stereo features derived from Siamese networks into differentiable geometric volumes. 
Further, \citet{wang2021plumenet} employ 3D occupancy to directly supervise the depth estimation model. The works in~\cite{guo2021liga, liu2023stereodistill} improve the performance by leveraging knowledge distillation. \citet{chen2022dsgn++} perform depth-wise reconstruction of the volume in geometric modeling to alleviate the bottleneck of 2D to 3D information propagation. These methods represent state-of-the-art (SOTA) for stereo-based 3D object detection, and our work explores the application of the work in~\cite{chen2022dsgn++} for streaming perception.

\paragraph{Streaming perception.}
The concept of streaming perception was initially introduced in \cite{li2020towards}, which evaluates streaming average precision (sAP) while accounting for latency considerations. With this metric, non-real-time detectors~\cite{lin2017focal, he2017mask} will lead to great performance degradation since they unavoidably miss some intermediate frames. 
Therefore, \citet{li2020towards} further propose to address this issue through decision-theoretic scheduling, asynchronous tracking \citep{bergmann2019tracking}, and future prediction \citep{kalman1960new}.

The work in~\cite{sela2022context} confirm the findings of~\cite{li2020towards}, and extend their analysis to object tracking. 
\citet{ghosh2023chanakya} propose a learned approximate execution framework to balance accuracy and latency implicitly. 
Instead of seeking better trade-offs or enhancing base detectors, 
StreamYOLO~\cite{yang2022real} simplifies streaming perception to a end-to-end task of ``predicting the next frame'' with a real-time detector. 
Based on this principle~\cite{yang2022real}, LongShortNet~\cite{li2023longshortnet} and DAMO-StreamNet~\cite{he2023damo} improve the streaming accuracy by leveraging longer temporal fusion and knowledge distillation respectively. 
The work in~\cite{wang2023we} expands the nuScenes dataset~\cite{caesar2020nuscenes} and introduces a 3D benchmark tailored for streaming perception assignments. To the best of our knowledge, there are currently no existing applications of 3D perception algorithms for this task.

\paragraph{Temporal 3D object detection.}
These methods can be broadly classified into three categories: 
(1) The \textit{LiDAR sequences-based methods}~\cite{yang20213d, yin2020lidar, yuan2021temporal, meyer2020laserflow} complement the 3D shape within a single frame by integrating historical features through temporal modeling. 
(2) The \textit{streaming data-based methods}~\cite{han2020streaming, frossard2021strobe, chen2021polarstream} treat each LiDAR packet as an independent sample without requiring a complete sweeping by the LiDAR. (3) The \textit{Video-based methods}~\cite{hu2019joint, brazil2020kinematic} extend the image-based methods by tracking and fusing the same objects across different frames. 
Note that the essential difference between streaming perception and these methods is the misalignment of ground truth and features.

\section{Our Approaches}
\subsection{Base detector}
\label{sec:base}
We choose the recently proposed DSGN++~\cite{chen2022dsgn++} as the base detector in our study. 
Due to its high latency, our first goal is to optimize its pipeline to ensure the completion of inference for the current frame before the arrival of the next frame.
	
\begin{figure*}[htbp]

	\centering
	\includegraphics[width=1\linewidth]{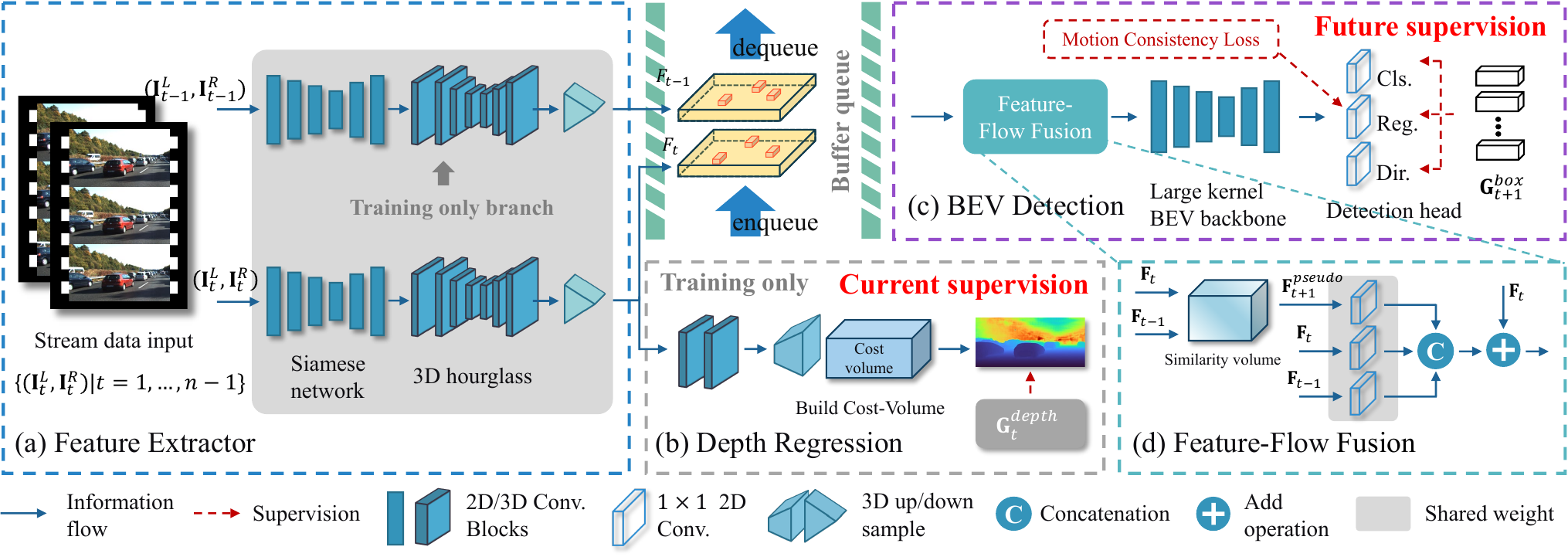}
	\caption{The architecture of StreamDSGN pipeline. (a) The feature extractor retrieves features from streaming stereo image pairs \( \{(\mathbf{I}^{L}_{t}, \mathbf{I}^{R}_{t})|t=1,...,n-1 \} \) and flattens them into BEV features. (b) The depth regression component utilizes \( \mathbf{G}^{depth}_{t} \) as the supervision. (c) The BEV detector predicts the object state of the next moment by merging features from the current and previous frames. (d) The Feature-Flow Fusion generates a pseudo-next feature \( \mathbf{F}^{pseudo}_{t+1} \) by extrapolating from past features and then concatenates it with the existing historical feature set \( \{ \mathbf{F}_{t}, \mathbf{F}_{t-1} \} \).}
	\label{fig:overview}

\end{figure*}

The optimization strategies are detailed in Appendix~\ref{app:opt}. These strategies aim to ensure real-time processing of the detector without significantly sacrificing the offline accuracy. Experiments in Appendix~\ref{app:opt_exp} show that the optimized model incurs only minor losses in offline accuracy while achieving a significant  reduction in latency, decreasing from 263.33ms to 80.71ms~(with a frame rate of 10Hz for the KITTI Tracking dataset~\cite{geiger2012we}).

Next, we extend the optimized model into a streaming data detection framework, as illustrated in Figure~\ref{fig:overview}.
During the training phase, we take the consecutive stereo images \( \left( \mathbf{I}^{L}_{t-1}, \mathbf{I}^{R}_{t-1}, \mathbf{I}^{L}_{t}, \mathbf{I}^{R}_{t} \right) \) as input and directly predict the result \( \mathbf{P}^{box}_{t+1} \) of the next frame. This data organization allows for random shuffling of training samples. 
Note that in streaming perception tasks, the input data cannot include time step \( t+1 \) since we cannot access future frames at the current moment in real-world scenarios.
During inference, as the input frames are continuous, we only need to input the current frame and store its intermediate feature in a buffer queue for fusion with the next input.


Different from previous 2D approaches~\cite{yang2022real, li2023longshortnet, he2023damo}, our temporal fusion is executed before the BEV backbone instead of preceding the detection head. This is because we believe that the shallow convolutional layers in the detection head lack an effective receptive field to extract large-span contextual information from the fused features. We validate this conjecture in the experiments presented in Appendix~\ref{app:fusion_stage}.

\begin{figure*}[htbp]
	\centering

	\includegraphics[width=0.92\linewidth]{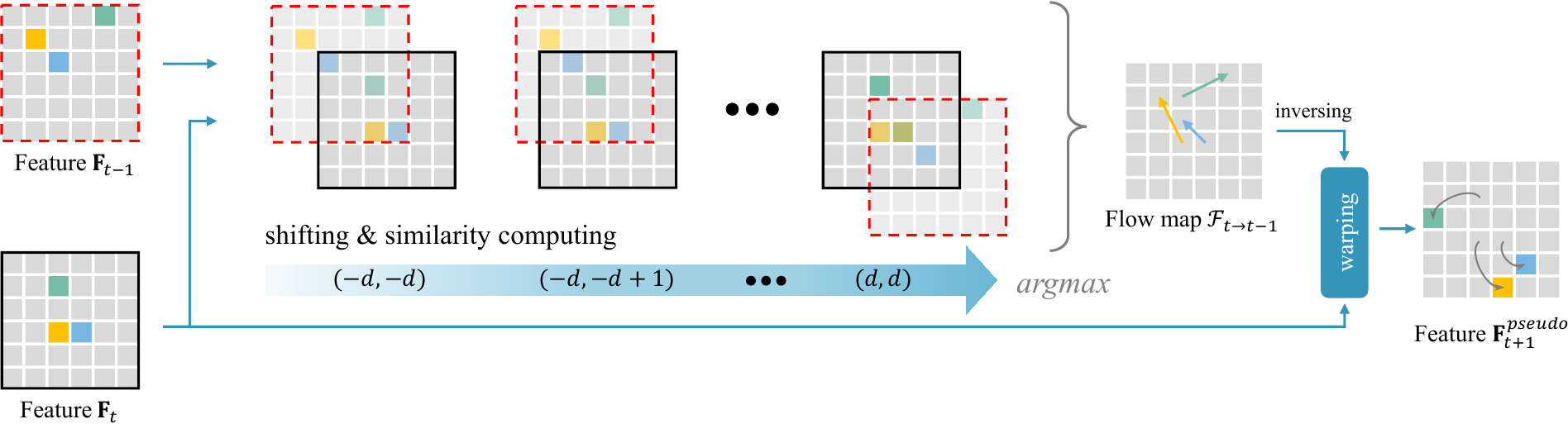}
	\caption{A toy example of pseudo-next feature generation. }
	\label{fig:feature_flow}

\end{figure*}

\subsection{Feature-Flow Fusion (FFF)}
\label{sec:fff}


Different from fusion methods for temporal 3D object detection~\cite{huang2022bevdet4d, li2022bevformer, liu2023petrv2, yang2023bevformer}, which aim to complement the geometric shape of objects, our Feature-Flow Fusion (FFF) is designed to warp current feature to align with the next ground truth. The pipeline of FFF is shown in Figure~\ref{fig:feature_flow}, it utilizes historical changes to accomplish the warping based on motion consistency.

\paragraph{Calculation of Feature Flow.}

%

Recall that our input features are \( \mathbf{F}_{t}, \mathbf{F}_{t-1} \in \mathbb{R}^{H\times W\times C} \). 
The first step of FFF involves computing the feature flow between the \( \mathbf{F}_{t} \) and \( \mathbf{F}_{t-1} \). This process is similar to the computation of similarity volume in optical flow~\cite{ilg2017flownet, sun2018pwc}.
We assume that the search space for the computing is discrete and rectangular. 
Specially, we define a shift indices set \( \mathcal{S} \) as: 
\begin{align}
	\mathcal{S} = \{ \left(-d,-d\right), \left(-d,-d+1\right), ..., (0, 0), ..., \left(d,d-1\right), \left(d,d\right) \}
\end{align}
where \( d \) is the maximal displacement. Then we shift the elements \( \{ \mathbf{c}_{t-1}(i,j)|i=0,...,H-1;j=0,...,W-1 \} \) in \( \mathbf{F}_{t-1} \) according to each shift index in \( \mathcal{S} \), and match the shifted \( \mathbf{F}_{t-1} \) with \( \mathbf{F}_{t} \) to compute the cosine similarity \( \mathbf{s}(i,j) \) of each matched elements. 
When considering a single match with a shift index of \( (\mathbf{u}_{s}, \mathbf{v}_{s}) \), the similarity at pixel index \( (\mathbf{u}, \mathbf{v}) \) can be defined as:
\begin{align}
	\mathbf{s}(\mathbf{u}, \mathbf{v}) &= \frac{1}{\| \mathbf{c}_{t-1}(\mathbf{u}+\mathbf{u}_{s}, \mathbf{v}+\mathbf{v}_{s})  \| \| \mathbf{c}_{t}(\mathbf{u}, \mathbf{v})  \|} \left( \left( \mathbf{c}_{t-1}(\mathbf{u}+\mathbf{u}_{s}, \mathbf{v}+\mathbf{v}_{s})\right)^{\intercal} \mathbf{c}_{t}(\mathbf{u}, \mathbf{v}) \right)
\end{align}
where \( \intercal \) represents the transpose operator. Stacking the whole similarity matrix \( \mathcal{C}=\{\mathbf{s}(i,j)|i=0,...,H-1;j=0,...,W-1 \} \) obtained from different shift indices along a new dimension, we can yield the similarity volume \( \mathcal{V}\in \mathbb{R}^{H\times W\times D} \), where \( D=\left(2d+1\right)^{2} \).

Next, we ascertain the feature flow \( \mathcal{F}_{t\rightarrow t-1}\in\mathbb{R}^{H\times W\times 2} \) between \( \mathbf{F}_{t} \) and \( \mathbf{F}_{t-1} \) by identifying the index with the maximum similarity. This process can be formulated as:
\begin{align}
	\mathcal{F}_{t\rightarrow t-1} = gather\left( \mathcal{S}, \underset{D}{\arg\max} \mathcal{V} \right)
\end{align}
Note that the acquisition of \( \mathcal{F}_{t\rightarrow t-1} \) does not require any learnable parameters. In our implementation, we accelerate the above process through parallelization, 
and reduce computation by downsampling \( \mathbf{F}_{t-1} \) and \( \mathbf{F}_{t} \) via max poooling,  
then restore the resolution of \( \mathcal{F}_{t\rightarrow t-1} \) through bilinear interpolation.


\paragraph{Pseudo-next Feature Generation.}
According to the motion consistency, with a sufficiently high frame rate, the magnitude of displacement of an object from \( t \) to \( t-1 \) is consistent with that from \( t \) to \( t+1 \), but in opposite directions, \textit{i.e.}, \( \mathcal{F}_{t\rightarrow t+1} \approx -\mathcal{F}_{t\rightarrow t-1} \). Thus, we can warp \( \mathbf{F}_{t} \) to the \( t+1 \) grid by using inverted \( \mathcal{F}_{t\rightarrow t-1} \). Let \( \mathbf{w}_{t \rightarrow t-1} \left(\mathbf{u}, \mathbf{v} \right) = \left( \mathbf{x}(\mathbf{u}, \mathbf{v}), \mathbf{y}(\mathbf{u}, \mathbf{v}) \right) \) denotes the flow at pixel index \( (\mathbf{u}, \mathbf{v}) \) in \( \mathcal{F}_{t\rightarrow t-1} \), 
where \( \mathbf{x} \) and \( \mathbf{y} \) represent values in the vertical and horizontal directions, respectively. 
Let \( \mathbf{F}^{pseudo}_{t+1}=\{\mathbf{c}^{pseudo}_{t+1}(i, j) | i=0,...,H-1;j=0,...,W-1 \} \) represents the warped feature in \( t+1 \) grid. The warped element at \( (\mathbf{u}, \mathbf{v}) \) can be formulated as~\cite{jaderberg2015spatial, ilg2017flownet, sun2018pwc}:
\begin{align}
	\mathbf{c}^{pseudo}_{t+1}(\mathbf{u}, \mathbf{v}) = \mathbf{c}_{t} \left( \mathbf{u} - \mathbf{x} \left( \mathbf{u}, \mathbf{v} \right), \mathbf{v} - \mathbf{y} \left( \mathbf{u}, \mathbf{v} \right) \right)
\end{align}
We refer to the \( \mathbf{F}^{pseudo}_{t+1} \) obtained through this pixel-level backward warping as pseudo-next feature. In theory, with a sufficiently fast frame rate and accurate matching, it can be aligned well with real \( \mathbf{F}_{t+1} \). As this process does not require the real \( \mathbf{F}_{t+1} \), 
it satisfies the constraints of streaming perception.

Note that \( \mathbf{F}^{pseudo}_{t+1} \) has limitations when dealing with occluded or truncated objects, and the warping operation may introduce additional edge noise into the scene. Thus, 
we fuse it with historical features to complement the geometric shape information of the objects. Similar to~\cite{yang2022real}, we initially utilize weight-shared convolutions to reduce the channels of \( \mathbf{F}_{t-1} \), \( \mathbf{F}_{t} \), and  \( \mathbf{F}^{pseudo}_{t+1} \), and then concatenate them together. The concatenated features are connected with \( \mathbf{F}_{t} \) via add operator to enhance the representation of static information, as illustrated in part (d) of Figure~\ref{fig:overview}.
The latency of our FFF process is only 7.67ms, please refer to Appendix~\ref{app:fff} for more latency and hyperparameter reports.

\subsection{Motion Consistency Loss (MCL)}

\begin{wrapfigure}{r}{0.5\textwidth}

	\centering
	\includegraphics[width=0.5\textwidth]{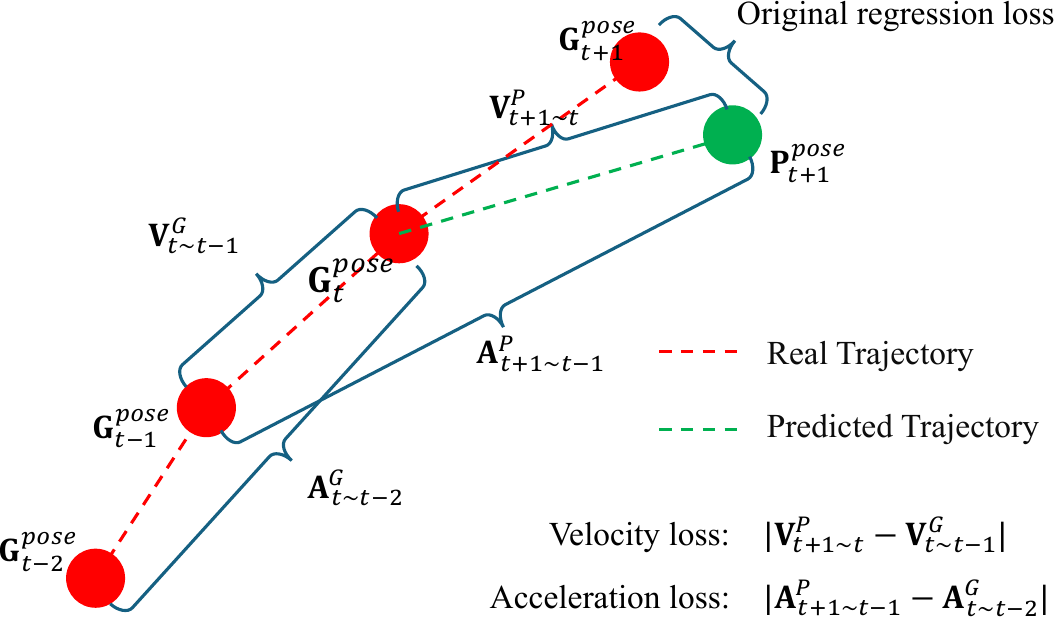}
	\caption{Illustration of MCL.}
	\label{fig:mcl}

\end{wrapfigure}

To explicitly guide the model in learning the offset magnitude, we propose a Motion Consistency Loss (MCL) as supplementary regression supervision, which consists of velocity loss and acceleration loss. This loss is also grounded on the principle of motion consistency. It leverages historical motion trajectories to guide the prediction of the next frame, as illustrated in Figure~\ref{fig:mcl}.


The initial step in calculating the MCL involves establishing correspondences between bounding boxes across different time steps. We establish these correspondences between the ground truth of \( \{ \mathbf{G}^{box}_{t-2}, \mathbf{G}^{box}_{t-1}, \mathbf{G}^{box}_{t} \} \) by utilizing target IDs. For the prediction \( \mathbf{P}^{box}_{t+1} \), we calculate an IoU (Intersection over Union) matrix with respect to \( \mathbf{G}^{box}_{t} \) and employ a \( \max \) operation to identify the index corresponding to the highest matching value, thus establishing their correspondence.

Let \( \mathbf{G}^{pose}_{i} \) denotes the ground truth center position and rotation angle of the objects at time step \( i \), where \( i=t-2, t-1, t \), such that \( \mathbf{G}^{pose}_{i}=(x^{g}_{i},y^{g}_{i},z^{g}_{i},\theta^{g}_{i}) \).
Similarly, \( \mathbf{P}^{pose}_{t+1} =(x^{p}_{t+1},y^{p}_{t+1},z^{p}_{t+1},\theta^{p}_{t+1}) \) represents the prediction for the next frame.
Base on the correspondence, now we can calculate the displacement vector and the sine difference of the rotation angles between \( \mathbf{P}^{pose}_{t+1} \) and \( \mathbf{G}^{pose}_{t} \) for the predicted offset \( \mathbf{V}^{p}_{t+1\rightarrow t}=(\Delta x^{p}, \Delta y^{p}, \Delta z^{p}, \Delta \theta^{p}) \):
\begin{align}
	\Delta x^{p} &= x^{p}_{t+1} - x^{g}_{t}, \qquad \Delta y^{p} = y^{p}_{t+1} - y^{g}_{t}, \nonumber \\
	\Delta z^{p} &= z^{p}_{t+1} - z^{g}_{t}, \qquad \Delta \theta^{p} = \sin(\theta^{p}_{t+1}-\theta^{g}_{t})
\end{align}
We can similarly calculate the ground truth offset \( \mathbf{V}^{g}_{t\rightarrow t-1} \) between \( \mathbf{G}^{pose}_{t} \) and \( \mathbf{G}^{pose}_{t-1} \) as supervision for the predicted offset \( \mathbf{V}^{p}_{t+1\rightarrow t} \). Thus, the regression for velocity loss can be formulated as:
\begin{align}
	\mathcal{L}_{\mathbf{V}} = SmoothL1(\mathbf{V}^{p}_{t+1\rightarrow t}-\mathbf{V}^{g}_{t\rightarrow t-1})
\end{align}
Further consideration is given to velocity change. We calculate the ground truth offset \( \mathbf{V}^{g}_{t-1 \rightarrow t-2} \) between \( \mathbf{G}^{pose}_{t-1} \) and \( \mathbf{G}^{pose}_{t-2} \) and then utilize the ground truth \( \mathbf{A}^{g}_{t \rightarrow t-2}=\mathbf{V}^{g}_{t \rightarrow t-1}-\mathbf{V}^{g}_{t-1 \rightarrow t-2} \) of velocity change to supervise the prediction \( \mathbf{A}^{p}_{t+1 \rightarrow t-1}=\mathbf{V}^{p}_{t+1 \rightarrow t}-\mathbf{V}^{g}_{t \rightarrow t-1} \). We also employ the Smooth L1 loss~\cite{yan2018second} to regress the acceleration consistency:
\begin{align}
	\mathcal{L}_{\mathbf{A}} = SmoothL1(\mathbf{A}^{p}_{t+1 \rightarrow t-1}-\mathbf{A}^{g}_{t \rightarrow t-2})
\end{align}
The MCL can then be defined as:
\begin{align}
	\mathcal{L}_{MCL} = \mathcal{L}_{\mathbf{V}} + \tau \mathcal{L}_{\mathbf{A}}
	\label{eq:tau}
\end{align}
where \( \tau=0.8 \). The grid search of \( \tau \) can be seen in Appendix~\ref{app:mcl}. Let \( \mathcal{L}_{ori} \) denotes the original loss of DSGN++ \cite{chen2022dsgn++} with the supervision of the next frame, our total loss is finally therefore:
\begin{align}
	\mathcal{L} = \frac{1}{N_{pos}}(\mathcal{L}_{ori}+\lambda\mathcal{L}_{MCL})
	\label{eq:lambda}
\end{align}
where \( N_{pos} \) is the number of positive anchors and \( \lambda=0.5 \).

\subsection{Large Kernel BEV Backbone (LKBB)}
\label{sec:lkbb}

\begin{figure*}[htbp]
	\centering
	\subcaptionbox{Architecture of a VAN block. \label{fig:lka}}{\includegraphics[width = 0.398\textwidth]{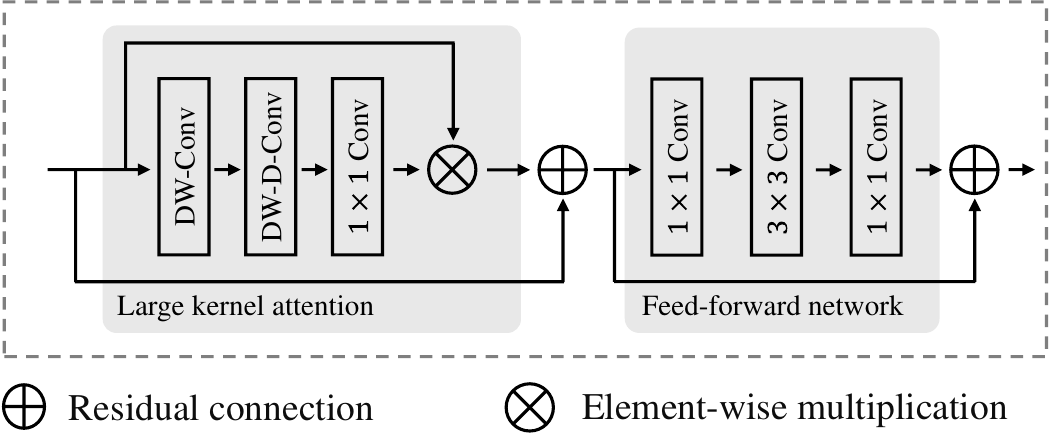}}
	\hfill
	\subcaptionbox{Architecture of LKBB. \label{fig:lkbb}}{\includegraphics[width = 0.58\textwidth]{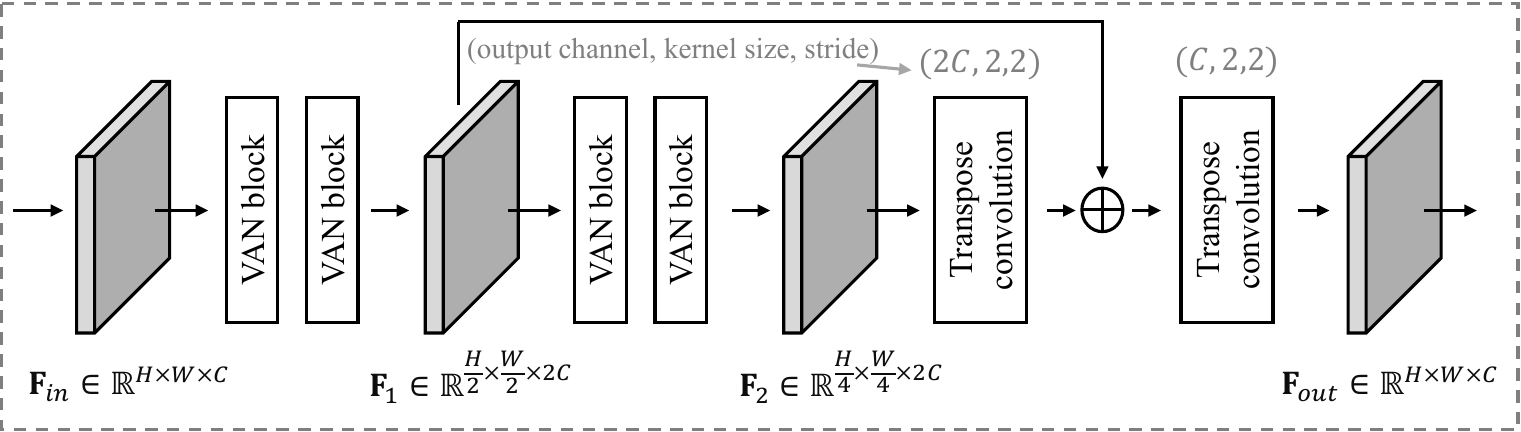}}
	\caption{Illustration of LKBB.}
	
\end{figure*}

Our LKBB is built upon Visual Attention Network (VAN)~\cite{guo2023visual}, which comprises a Large Kernel Attention (LKA)~\cite{guo2023visual} module and a Feed-Forward Network (FFN)~\cite{wang2022pvt}, as shown in the Figure~\ref{fig:lka}. 
The LKA module is composed of cascaded depth-wise convolution (DW-Conv), depth-wise dilation convolution (DW-D-Conv), and \( 1\times1 \) convolution.



We migrate VAN into our BEV backbone network to increase its receptive field, as illustrated in Figure~\ref{fig:lkbb}. For fair comparison, we adjust the number of VAN blocks and the multiplier for channel transformations to keep the parameter count of LKBB similar to that of the original structure. Further, to maintain the sensitivity of the model to fine-grained geometric details, we incorporate multi-scale features via residual connections. 


Let \( f^{c, k, s}_{\intercal} \) denotes the transpose convolution operation, where \( c \), \( k \) and \( s \) respectively represent the number of filters, kernel size and stride. The multi-scale fusion process can be formulated as:
\begin{align}
	\mathbf{F}_{out} = f^{C, 2, 2}_{\intercal}  \left(f^{2C, 2, 2}_{\intercal}\left( \mathbf{F}_{2} \right) + \mathbf{F}_{1} \right)
	\label{eq:lkbb}
\end{align}
where \( \mathbf{F}_{1} \), \( \mathbf{F}_{2} \) and \( \mathbf{F}_{out} \) represent the downsampled features from two stages and the final output feature, respectively, with dimensions\( \frac{H}{2} \times \frac{W}{2} \times 2C \), \( \frac{H}{4} \times \frac{W}{4} \times 2C \) and \( H \times W \times C \). 

We report the comparison of parameter count and computational complexity between our LKBB and the original hourglass backbone~\cite{newell2016stacked} in Appendix~\ref{app:lkbb}. Specifically, our parameter count is similar to the original structure while the computational complexity is reduced by approximately 6GFLOPs.

\section{Experiments}
\label{sec:exp}

\subsection{Experimental setup}

\paragraph{Dataset.}
The prevalent autonomous driving datasets, such as nuScenes~\cite{caesar2020nuscenes}, nuScenes-H~\cite{wang2023we} and Waymo Open~\cite{sun2020scalability}, lack stereo image provision. Further, the stereo frame rate (only 5Hz) within Argoverse~\cite{chang2019argoverse} is insufficient for streaming simulation. Consequently, we conduct our experiments on KITTI tracking dataset~\cite{geiger2012we}, which provides stereo images and a higher frame rate (10Hz). We partition the training set with 20 scenes into numerous frame sequences, each comprising 40 frames. Among them, the even-numbered sequences are used for training with a total of 4,291 frames, and the odd-numbered sequences are used for testing with a total of 3,672 frames. 
We analyze the domain gap of this partitioning method in Appendix~\ref{app:gap}.

\paragraph{Evaluation metrics.}
We follow~\cite{li2020towards,wang2023we} to conduct the streaming simulation to obtain the streaming average precision (sAP) for both BEV and 3D perspectives.
Consistent with KITTI~\cite{geiger2012we}, objects are categorized into three levels based on their recognition complexity: Easy, Moderate, and Hard. All of our precision measurements for experiments and ablation studies are calculated at \( IoU=0.7 \) and with 40 recall positions. If not specified, all results are for the ``Car'' category.

\paragraph{Implementation details.}
Our experiments are conducted on the NVIDIA TITAN RTX platform with a total of 20 epochs. During the training phase, the Adam optimizer~\cite{kingma2014adam} is employed in conjunction with the OneCycle learning rate decay strategy~\cite{smith2019super}. The initial learning rate is set to \( 1e-3 \), progressively increased to \( 1e-2 \), and finally decayed to \( 1e-8 \). Further, for a fair comparison with offline perception, we extend the copy-paste data augmentation~\cite{yan2018second, chen2022dsgn++} to streaming data. It cascades moving foreground objects into consecutive training frames to enrich the training samples.

\begin{table}
	\centering
	\caption{Comparison with the meta-detector named Streamer. DSGN++$_{opt}$ represents the method equipped with our real-time optimization strategies. The symbol ``$\dagger$'' denotes the basic framework of StreamYOLO~\cite{yang2022real}, which is built upon the DSGN++$_{opt}$ architecture. }
	\label{tab:main}
	\resizebox{1\linewidth}{!}{
		\begin{tblr}{colspec=c|c|c|cccccc|cccccc, stretch=1., rows={ht=1.25\baselineskip}}
			\hline[1.25pt]
			\SetCell[r=2]{c}{Methods} & \SetCell[r=2]{c}{Base detector} &\SetCell[r=2]{c}{Latency \\ (ms)} & \SetCell[c=3]{c} sAP$\rm _{BEV}$ (IoU=0.5) && & \SetCell[c=3]{c} sAP$\rm _{3D}$ (IoU=0.5) && & \SetCell[c=3]{c}sAP$\rm _{BEV}$ (IoU=0.7) && & \SetCell[c=3]{c}sAP$\rm _{3D}$ (IoU=0.7) && \\
			& & & Easy & Mod. & Hard & Easy & Mod. & Hard & Easy & Mod. & Hard & Easy & Mod. & Hard \\
			\hline
			\hline
			\SetCell[c=14]{c} \textbf{Car} \\
			\hline
			Streamer & DSGN++~ & 263.33 & 14.76 & 11.56 & 10.62 & 12.51 & 8.86 & 7.74 & 5.26 & 3.42 & 3.11 & 2.82 & 1.92 & 1.51 \\
			Streamer & DSGN++$_{opt}$ & 80.71 & 70.57 & 61.04 & 56.31 & 64.47 & 55.18 & 50.36 & 30.89 & 22.50 & 19.96 & 25.50 & 17.68 & 14.75 \\ 
			StreamYOLO$^\dagger$ & DSGN++$_{opt}$ & 81.32 & 92.22 & 85.75 & 82.45 & 89.40 & 82.89 & 78.10 & 80.72 & 68.51 & 63.14 & 73.02 & 58.37 & 51.86 \\
			StreamDSGN (ours) & DSGN++$_{opt}$ & 91.45  & 93.10 & 87.46 & 84.34 & 92.53 & 84.55 & 81.34 & 85.40 & 72.47 & 68.66 & 77.47 & 63.76 & 57.42 \\
			\hline
			\SetCell[c=14]{c} \textbf{Pedestrian} \\
			\hline
			Streamer & DSGN++ & 263.33 & 7.38 & 7.22 & 7.42 & 6.58 & 6.85 & 6.59 & 1.87 & 1.68 & 1.51 & 1.55 & 1.13 & 1.12 \\
			Streamer & DSGN++$_{opt}$ & 80.71 & 58.23 & 54.57 & 50.25 & 56.56 & 53.97 & 49.70 & 31.26 & 28.92 & 26.55 & 25.18 & 23.50 & 21.49 \\ 
			StreamYOLO$^\dagger$ & DSGN++$_{opt}$ & 81.32 & 74.54 & 69.10 & 63.48 & 73.44 & 68.68 & 63.08 & 53.27 & 49.35 & 44.67 & 45.53 & 42.15 & 38.01 \\
			StreamDSGN (ours) & DSGN++$_{opt}$ & 91.45 & 80.70 & 75.38 & 69.23 & 80.41 & 75.12 & 68.96 & 62.33 & 57.51 & 51.34 & 54.12 & 50.05 & 44.89 \\
			\hline
			\SetCell[c=14]{c} \textbf{Cyclist} \\
			\hline
			Streamer & DSGN++ & 263.33 & 3.73 & 4.29 & 4.31 & 3.74 & 4.15 & 3.74 & 1.13 & 0.83 & 0.82 & 1.13 & 0.60 & 0.60 \\
			Streamer & DSGN++$_{opt}$ & 80.71 & 41.11 & 40.73 & 40.07 & 40.78 & 40.37 & 39.24 & 7.56 & 15.52 & 14.82 & 6.77 & 14.35 & 14.13 \\ 
			StreamYOLO$^\dagger$ & DSGN++$_{opt}$ & 81.32 & 39.34 & 41.31 & 40.59 & 38.72 & 40.20 & 39.59 & 31.37 & 34.03 & 33.43 & 30.63 & 33.14 & 32.36 \\
			StreamDSGN (ours) & DSGN++$_{opt}$ & 91.45 & 44.62 & 48.92 & 48.74 & 43.52 & 48.03 & 47.46 & 37.42 & 41.10 & 40.64 & 35.01 & 37.37 & 37.17 \\
			\hline[1.25pt]
		\end{tblr}
	}

\end{table}

\begin{table}[htbp]
	\centering

	\caption{Ablation studies of our methods. Setting \textit{\textbf{a}} denotes the DSGN++$_{opt}$. Setting \textit{\textbf{b}} represents directly predicting future results. Setting \textit{\textbf{c}} incorporates historical feature based on \textit{\textbf{b}} and serves as our baseline method (highlighted in \textcolor{gray}{gray}). Settings \textit{\textbf{d}}, \textit{\textbf{e}}, and \textit{\textbf{f}} respectively incorporate our enhancement strategies. Setting \textit{\textbf{g}} is our final proposed solution (highlighted in \textcolor{green}{green}).}
	\label{tab:main_abla}
	\resizebox{1\linewidth}{!}{
		\begin{tblr}{colspec=c|cc|ccc||ccc|ccc, stretch=1., rows={ht=1.25\baselineskip}}
			\hline[1.25pt]
			\SetCell[r=2]{c} {Setting ID} & \SetCell[c=2]{c} {Pipeline} & & \SetCell[r=2]{c} {FFF} & \SetCell[r=2]{c} {MCL} & \SetCell[r=2]{c} {LKBB} & \SetCell[c=3]{c} sAP$\rm _{BEV}$ & & & \SetCell[c=3]{c} sAP$\rm _{3D}$ & & \\
			\cline{2-3}
			& predict \( t+1 \) & fuse \( t-1 \) & & & & Easy & Mod. & Hard & Easy & Mod. & Hard \\
			\hline
			\hline
			\textit{\textbf{a}} & $-$ & $-$ & $-$ & $-$ & $-$ & 30.89 & 22.50 & 19.96 & 25.50 & 17.68 & 14.75 \\
			\textit{\textbf{b}} & $\checkmark$ & $-$ & $-$ & $-$ & $-$ & 53.73  & 47.46  & 43.36  & 38.62  & 32.55  & 29.39 \\  
			\SetRow{bg=gray!25}
			\textit{\textbf{c}} & $\checkmark$ & $\checkmark$ & $-$ & $-$ & $-$ & 83.20 & 71.15 & 65.74 & 75.38 & 59.55 & 53.11 \\
			\textit{\textbf{d}} & $\checkmark$ & $\checkmark$ & $\checkmark$ & $-$ & $-$ & 85.08 & 71.98 & 66.49 & 75.88 & 61.57 & 54.77\\
			\textit{\textbf{e}} & $\checkmark$ & $\checkmark$ & $-$ & $\checkmark$ & $-$ & 84.63 & 71.01 & 66.18 & 78.15 & 62.28 & 56.80\\
			\textit{\textbf{f}} & $\checkmark$ & $\checkmark$ & $-$ & $-$ & $\checkmark$ & 84.97 & 71.96 & 66.53 & 76.74 & 60.31 & 55.01 \\  
			\SetRow{bg=green!25}
			\textit{\textbf{g}} & $\checkmark$ & $\checkmark$ & $\checkmark$ & $\checkmark$ & $\checkmark$ & 85.40 & 72.47 & 68.66 & 77.47 & 63.76 & 57.42 \\
			
			\hline[1.25pt]
		\end{tblr}	
	}

\end{table}



\subsection{Comparison with Meta-detector}

We re-implement the meta-detector named Streamer for streaming benchmark in the stereo-based 3D object detection domain. 
Following the works in~\cite{li2020towards, wang2023we}, Streamer respectively employs DSGN++~\cite{chen2022dsgn++} and DSGN++$_{opt}$ (equipped with real-time optimization strategies) as the base detector to predict the current frame, and then utilizes a Kalman filter~\cite{kalman1960new} to combine historical outputs for forecasting the future state. 
Note that Streamer~\cite{li2020towards, wang2023we} is a non-end-to-end solution.
In Table~\ref{tab:main}, we compare our method with these non-real-time and real-time settings. 
From these comparisons, we can make the following observations.

First, the Streamer~\cite{li2020towards, wang2023we} combined with DSGN++~\cite{chen2022dsgn++} setting results in a noticeable inferior in streaming accuracy. In our streaming simulation, we find that this is due to the large-span temporal interval, making it challenging for the Kalman filter~\cite{kalman1960new} to accurately track each target in multi-object scenarios. Second, the Streamer~\cite{li2020towards, wang2023we} combined with the DSGN++$_{opt}$ setting shows higher performance compared to the DSGN++ combined one due to the latency reduction. However, with a high matching threshold (\( IoU=0.7 \)), the improvement in streaming accuracy is quite limited. Third, the StreamYOLO~\cite{yang2022real} framework built upon DSGN++$_{opt}$ achieves significant improvements across all metrics. These results demonstrate the competitiveness of the end-to-end framework that directly predicts future states compared to non-end-to-end solutions in the benchmark. Finally, our StreamDSGN reaches the highest level of performance, with approximately a 5\% improvement in each metric compared to the extended version of StreamYOLO~\cite{yang2022real}. Despite the latency in this setting reaching 91.45ms, the inference speed remains faster than the input frame rate, thus meeting real-time requirements.

\begin{figure*}[htbp]

	\centering
	\includegraphics[width=1\linewidth]{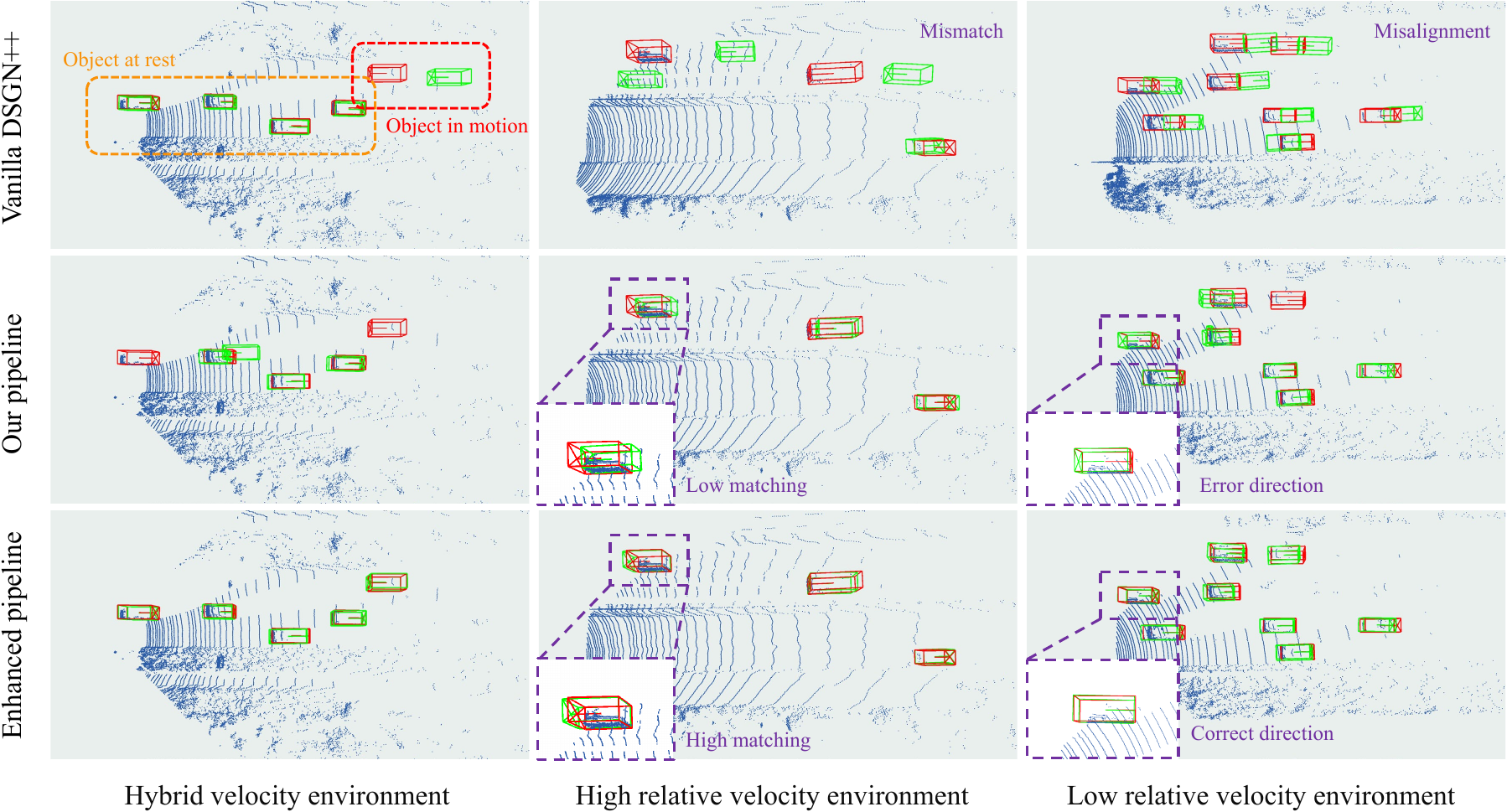}
	\caption{Qualitative analysis of different relative velocity scenarios. We visualize the predictions in point clouds for a clearer comparison. The first row describes DSGN++ \cite{chen2022dsgn++} without any modifications. The second row integrates real-time optimization and fusion of historical frames to predict the next frame. The third row showcases our three enhancement strategies. Ground truth and prediction instances are respectively delineated by \textcolor{red}{red} and \textcolor{green}{green} bounding boxes, with lines inside the boxes indicating the orientation of objects.}
	\label{fig:visual}

\end{figure*}

\subsection{Ablation Studies}

To validate the effectiveness of each strategy, we conduct ablation experiments by toggling them on and off. Experimental results is illustrated in Table~\ref{tab:main_abla}. 
In the following, we provide a more detailed description of the ablation experiments. 

\paragraph{Effectiveness of the Pipeline.}
The comparison between setting \textit{\textbf{b}} and setting \textit{\textbf{a}} reveals that directly predicting future states by using a real-time detector can alleviate the accuracy degradation caused by streaming perception constraints. However, the improvement in accuracy at this time is limited due to the lack of information on each object's displacement magnitude. Furthermore, by incorporating historical features, streaming accuracy can be significantly improved (see setting \textit{\textbf{c}}). This improvement occurs because the fused features implicitly embed motion cues, enabling the model to learn different offsets for each object. Thus, setting \textit{\textbf{c}} serves as our baseline and is utilized for subsequent ablation studies to assess each enhancement strategy's effectiveness.

\paragraph{Effectiveness of Enhanced Strategies.} The comparison of settings \textit{\textbf{d}}, \textit{\textbf{e}}, and \textit{\textbf{f}} with setting \textit{\textbf{c}} demonstrates the effectiveness of each enhancement strategy, respectively.
From these comparisons, we can make the following observations. 

First, since the fusion scheme used in setting \textit{\textbf{c}} is derived from the Dual-Flow module of StreamYOLO~\cite{yang2022real}, the comparison between setting \textit{\textbf{d}} and setting \textit{\textbf{c}} can be viewed as a comparison between our FFF and the SOTA streaming perception fusion method. The results show that our FFF achieves improvements across all metrics. Second, setting \textit{\textbf{e}} reveals the effectiveness of MCL, as the additional supervision of trajectory consistency yields significant improvements. Compared to the baseline, improvements across the three levels reach 2.77\%, 2.73\%, and 3.69\% in sAP$\rm _{3D}$, respectively. Third, due to the consistency between the downsampling dimensions of the feature maps in LKBB and the baseline, as well as the similar parameter counts between the two structures, it is evident that the improvements in setting \textit{\textbf{f}} are primarily attributed to the advantage of the large receptive field feature extractor in capturing long-range contextual features. Finally, when we combine all strategies together (highlighted in \textcolor{green}{green}), the improvement in streaming accuracy reaches its peak, with a 4.33\% increase in sAP$\rm _{3D}$ compared to the baseline at the hard level.

\subsection{Qualitative Analysis}

\paragraph{Visualization of Streaming Detection.}
Figure~\ref{fig:visual} presents the qualitative results of our proposed pipeline compared to the vanilla DSGN++~\cite{chen2022dsgn++}. In scenarios where surrounding vehicles remain relatively stationary, DSGN++~\cite{chen2022dsgn++} demonstrates a robust capacity for accurately aligning with ground truth. However, as these vehicles initiate motion relative to the ego vehicle, DSGN++~\cite{chen2022dsgn++} exhibits misalignments or mismatches in predictions due to the latency. 
In contrast, our baseline pipeline adeptly addresses this issue. With the enhancement strategy incorporating FFF, MCL and LKBB, the alignment between predicted boxes and ground truth boxes is further improved.

\begin{figure*}
	\centering
	\subcaptionbox{Previous feature}{\includegraphics[width = 0.235\textwidth]{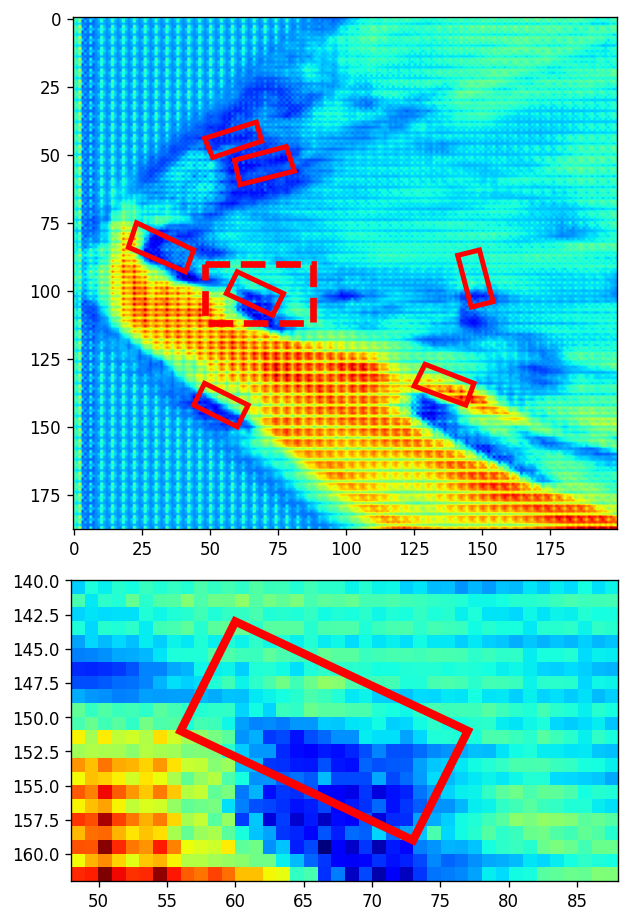}}
	\hspace{0.1em}
	\subcaptionbox{Current feature}{\includegraphics[width = 0.235\textwidth]{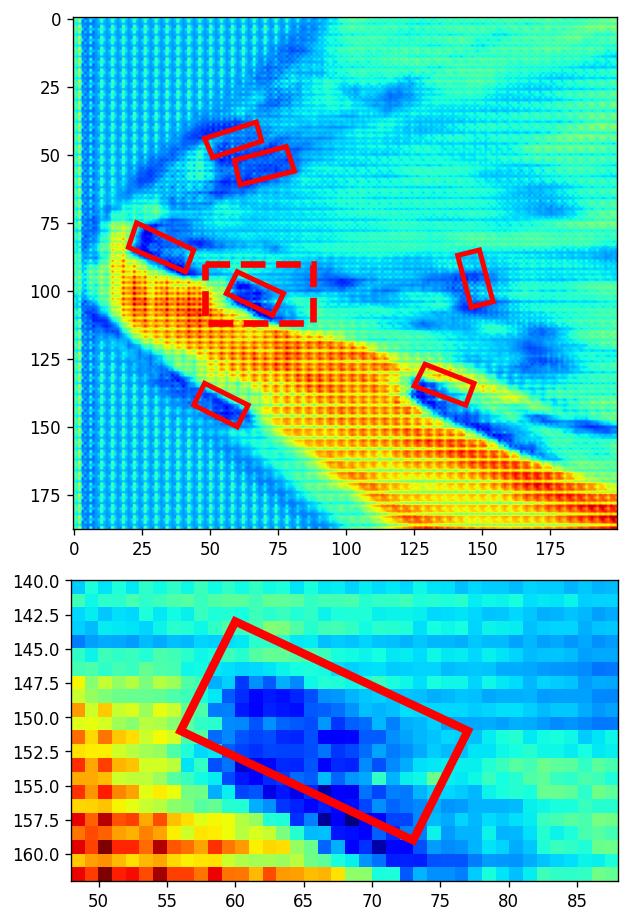}}
	\hspace{0.1em}
	\subcaptionbox{Next feature}{\includegraphics[width = 0.235\textwidth]{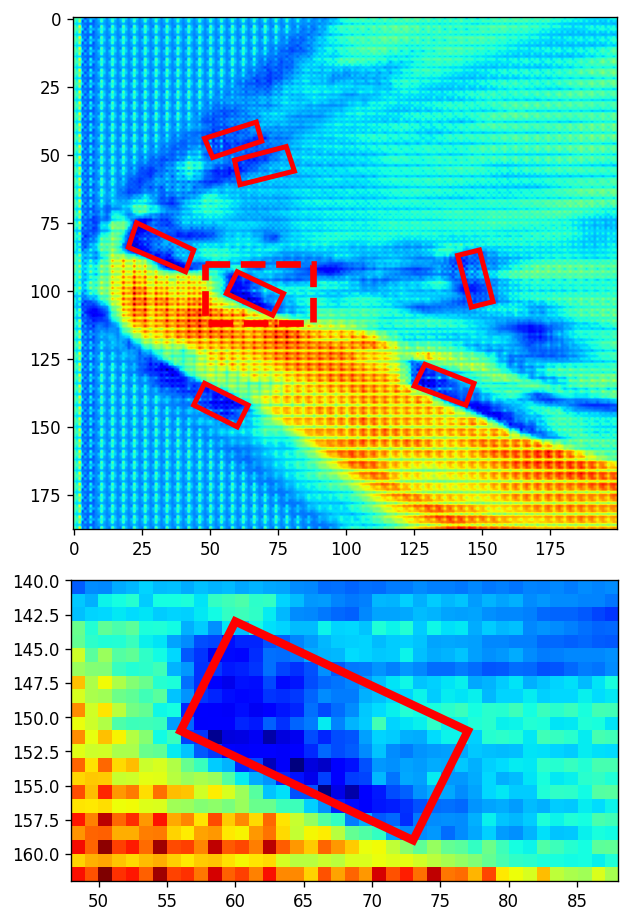}}
	\hspace{0.1em}
	\subcaptionbox{Pseudo-next feature}{\includegraphics[width = 0.235\textwidth]{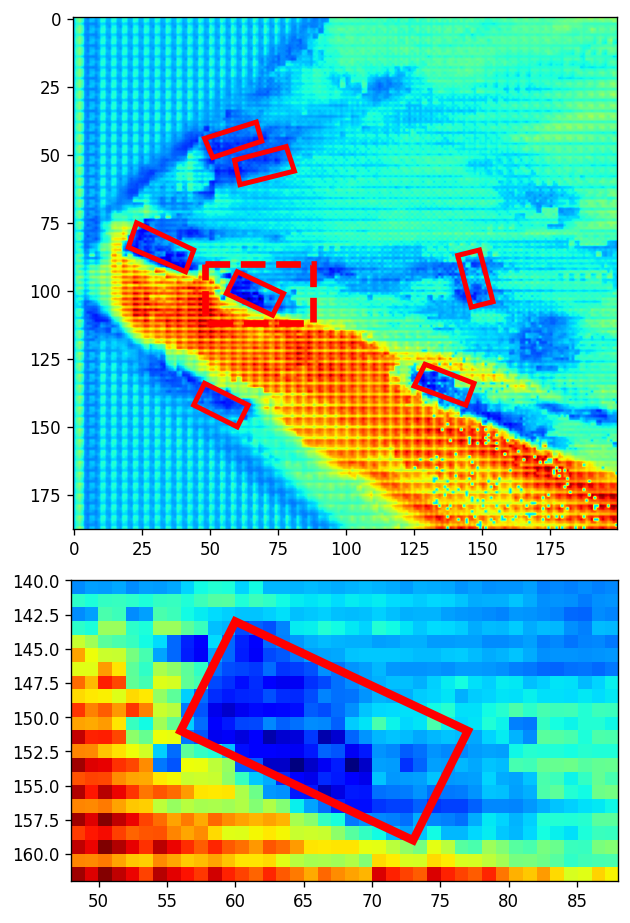}}  
	\caption{Qualitative analysis of the pseudo-next feature. The first row displays complete feature maps from different time steps, while the second row shows corresponding local regions of the feature maps. The \textbf{solid} red box and \textbf{dashed} box respectively represent the ground truth of the next frame used for supervision and the locally zoomed-in area.}
	\label{fig:visualized_warp}

\end{figure*}

\paragraph{Visualization of Pseudo-next Feature.}
We visualize the pseudo-next feature in Figure~\ref{fig:visualized_warp} for qualitative analysis. We can observe that due to the constraints of streaming perception, there is varying misalignment between the previous feature and the current feature with respect to the next ground truth used for supervision. However, after similarity matching and reverse warping, our pseudo-next feature aligns well with the ground truth.

\section{Conclusions}
\label{sec:con}

For the first time, we propose StreamDSGN, a real-time stereo-based 3D object detection framework for streaming perception. 
It is further equipped with Feature Flow Fusion, Motion Consistency Loss, and a Large Kernel BEV Backbone to enhance performance. 

\paragraph{Limitations and Future Work.}
When objects are occluded or truncated, FFF may produce incorrect pseudo-next features due to erroneous similarity matching. Currently, we mitigate this issue by simply integrating historical features. 
In the future, we plan to leverage neural networks to directly predict the flow of dynamic foreground objects, aiming to be more robust in addressing this challenge by exploiting the adaptability of neural networks. 
The second limitation of our approach is that it has only been validated using stereo-based methods. In contrast, currently mainstream autonomous driving perception tasks typically employ surrounding multi-view camera systems, which differ from stereo-based methods in their construction of BEV representations. Consequently, our future research will focus on validating the effectiveness of this approach within the context of multi-view methods.


\section{Acknowledgments}
This research was supported by the National Natural Science Foundation of China under Grant 62027804, the Fund of National Key Laboratory of Multispectral Information Intelligent Processing Technology (No. 202410487201), and the Major Key Project of PCL (PCL2021A13).

\label{ack}
	
	\bibliographystyle{plainnat}
	\bibliography{Ref}
	
	\newpage
\section*{NeurIPS Paper Checklist}

\begin{enumerate}
	
	\item {\bf Claims}
	\item[] Question: Do the main claims made in the abstract and introduction accurately reflect the paper's contributions and scope?
	\item[] Answer: \answerYes{} 
	\item[] Justification: We delineate our motivations, claims, and contributions in the abstract and introduction sections. Please see Section~\ref{sec:intro}
	
	\item {\bf Limitations}
	\item[] Question: Does the paper discuss the limitations of the work performed by the authors?
	\item[] Answer: \answerYes{} 
	\item[] Justification: Please refer to Section~\ref{sec:con} for a more detailed discussion on limitations.
	
	\item {\bf Theory Assumptions and Proofs}
	\item[] Question: For each theoretical result, does the paper provide the full set of assumptions and a complete (and correct) proof?
	\item[] Answer: \answerNA{} 
	\item[] Justification: Our work does not encompass theoretical results.
	
	\item {\bf Experimental Result Reproducibility}
	\item[] Question: Does the paper fully disclose all the information needed to reproduce the main experimental results of the paper to the extent that it affects the main claims and/or conclusions of the paper (regardless of whether the code and data are provided or not)?
	\item[] Answer: \answerYes{} 
	\item[] Justification: We have provided detailed descriptions of the experimental setup for reproducibility, including dataset configurations and model hyperparameters. For further details, please refer to Section~\ref{sec:exp}.

	\item {\bf Open access to data and code}
	\item[] Question: Does the paper provide open access to the data and code, with sufficient instructions to faithfully reproduce the main experimental results, as described in supplemental material?
	\item[] Answer: \answerYes{} 
	\item[] Justification: Our code is temporarily hosted on an anonymous platform: \href{https://anonymous.4open.science/r/streamDSGN-FD29}{\url{https://anonymous.4open.science/r/streamDSGN-FD29}.}. If the paper is accepted, we will release the source code on GitHub.

	\item {\bf Experimental Setting/Details}
	\item[] Question: Does the paper specify all the training and test details (e.g., data splits, hyperparameters, how they were chosen, type of optimizer, etc.) necessary to understand the results?
	\item[] Answer: \answerYes{} 
	\item[] Justification: Please refer to Section~\ref{sec:exp} for a more detailed discussion of the experimental setup.
	
	\item {\bf Experiment Statistical Significance}
	\item[] Question: Does the paper report error bars suitably and correctly defined or other appropriate information about the statistical significance of the experiments?
	\item[] Answer: \answerYes{} 
	\item[] Justification: Our experiments do not include error bars or confidence intervals, but they support the main claims of the paper.
	
	\item {\bf Experiments Compute Resources}
	\item[] Question: For each experiment, does the paper provide sufficient information on the computer resources (type of compute workers, memory, time of execution) needed to reproduce the experiments?
	\item[] Answer: \answerYes{} 
	\item[] Justification: We introduce the computational resources in the experimental setup. Please refer to Section~\ref{sec:exp} for details.
	
	\item {\bf Code Of Ethics}
	\item[] Question: Does the research conducted in the paper conform, in every respect, with the NeurIPS Code of Ethics \url{https://neurips.cc/public/EthicsGuidelines}?
	\item[] Answer: \answerYes{} 
	\item[] Justification: Our research adheres to the ethical guidelines of NeurIPS.

	\item {\bf Broader Impacts}
	\item[] Question: Does the paper discuss both potential positive societal impacts and negative societal impacts of the work performed?
	\item[] Answer: \answerNA{} 
	\item[] Justification: Our work has no societal impact.
	
	\item {\bf Safeguards}
	\item[] Question: Does the paper describe safeguards that have been put in place for responsible release of data or models that have a high risk for misuse (e.g., pretrained language models, image generators, or scraped datasets)?
	\item[] Answer: \answerNA{} 
	\item[] Justification: Our work does not entail such risks.
	
	\item {\bf Licenses for existing assets}
	\item[] Question: Are the creators or original owners of assets (e.g., code, data, models), used in the paper, properly credited and are the license and terms of use explicitly mentioned and properly respected?
	\item[] Answer: \answerYes{} 
	\item[] Justification: Our code is based on several released works. All have been cited.
	
	\item {\bf New Assets}
	\item[] Question: Are new assets introduced in the paper well documented and is the documentation provided alongside the assets?
	\item[] Answer: \answerYes{} 
	\item[] Justification: We have released an anonymous version of the code.
	
	\item {\bf Crowdsourcing and Research with Human Subjects}
	\item[] Question: For crowdsourcing experiments and research with human subjects, does the paper include the full text of instructions given to participants and screenshots, if applicable, as well as details about compensation (if any)? 
	\item[] Answer: \answerNA{} 
	\item[] Justification: Our work does not involve crowdsourcing or research with human subjects.
	
	\item {\bf Institutional Review Board (IRB) Approvals or Equivalent for Research with Human Subjects}
	\item[] Question: Does the paper describe potential risks incurred by study participants, whether such risks were disclosed to the subjects, and whether Institutional Review Board (IRB) approvals (or an equivalent approval/review based on the requirements of your country or institution) were obtained?
	\item[] Answer: \answerNA{} 
	\item[] Justification: Our work does not involve crowdsourcing or research with human subjects.
	
\end{enumerate}

	\vspace*{3em}

\appendix

\section{Appendix / supplemental material}
\subsection{Real-time Optimization}
\label{app:opt}
Our real-time optimization strategies are outlined as follows:


\begin{itemize}[leftmargin=3em]
	\item[(1)] We replace the original ResNet34 with ResNet18~\cite{he2016deep} as the backbone for stereo images.
	\item[(2)] We abandon the Dual-view Stereo Volume representation in DSGN++~\cite{chen2022dsgn++} and instead retained only one of PSV or 3DGV. See more detail in~\cite{chen2022dsgn++}.
	\item[(3)] Inspired by~\cite{lang2019pointpillars, li2023pillarnext}, we extend the grid size of the feature volume from \( (0.2m, 0.2m, 0.2m) \) to \( (0.2m, 0.2m, 0.4m) \) along the vertical axis.
	\item[(4)] We perform edge cropping for detection range~\cite{ma2021delving}. The new range is set to  \( [2m, 53.2m] \) for the Z-axis (depth), \( [-28.8m, 28.8m] \) for the X-axis, and \( [-1m, 3m] \) for the Y-axis respectively (in the camera coordinate system). 
	\item[(5)] Referring to~\cite{micikevicius2018mixed}, we incorporate Automatic Mixed Precision (AMP) techniques during both the training and inference stages.
\end{itemize}


\subsection{Effectiveness of Real-Time Optimization} 
\label{app:opt_exp}
\begin{table}[htbp]
	\centering
	\begin{minipage}[t]{1\textwidth}
	\vspace*{-1em}
	\centering
	\caption{Comparison of real-time optimization strategies. The term ``r18'' denotes the use of ResNet18 as the image feature extractor. ``Only 3DGV'' and ``Only PSV'' represent the exclusive adoption of 3DGV and PSV as stereo features, respectively. ``Low-res. (h)'' indicates low-resolution representation in the vertical dimension. ``EC'' signifies edge cropping, and ``AMP'' stands for the utilization of automatic mixed precision \cite{micikevicius2018mixed} during both training and inference stages. }
	\label{tab:opt}%
	\resizebox{0.8\linewidth}{!}{
		\begin{tblr}{colspec=cccccc|c|ccc|ccc, stretch=1.,  rows={ht=1.25\baselineskip}}
			\hline[1.25pt]
			\SetCell[c=6]{c} Ablation method & & & & & & \SetCell[r=2]{c}{ Latency \\ (ms)} & \SetCell[c=3]{c} offAP$\rm _{3D}$ & & & \SetCell[c=3]{c} sAP$\rm _{3D}$ & & \\ 
			\cline{1-6} \cline{8-13}
			r18 & Only 3DGV & only PSV & Lo-res. (h) & EC & AMP & & Easy & Mod. & Hard & Easy & Mod. & Hard \\
			\hline 
			\hline
			\SetCell[c=13]{c} Non-real-time (KF Forecasting \cite{kalman1960new, li2020towards, wang2023we}) \\
			\hline
			$-$     & $-$     & $-$     &  $-$     & $-$    & $-$    & 263.33  & 91.79  & 75.35  & 69.79  & 2.82  & 1.92  & 1.51  \\
			$\checkmark$	& $\checkmark$ &  $-$     &   $-$      &  $-$  & $-$    & 180.96  & 90.60  & 74.08  & 68.35  & 6.46  & 4.26  & 3.58  \\
			$\checkmark$ & $\checkmark$      &  $-$     &  $\checkmark$         &  $-$  & $-$    & 150.55  & 90.77  & 74.45  & 68.73  & 8.94  & 5.80  & 4.74  \\
			$\checkmark$ & $-$    &  $\checkmark$     &  $\checkmark$         &   $-$  & $-$   & 147.81  & \textbf{91.98}  & 75.91  & 68.96  & 7.51  & 5.15  & 4.41  \\
			$\checkmark$ & $-$ & $\checkmark$ & $\checkmark$  & $\checkmark$  & $-$  & 134.21 & 91.81 & \textbf{77.22} & \textbf{69.98} & 9.52 & 6.08 & 5.15 \\
			\hline
			\SetCell[c=13]{c} Real-time (End-to-end) \\
			\hline
			$\checkmark$ & $-$ & $\checkmark$ & $\checkmark$ & $-$  & $\checkmark$  & 86.68 & 91.67 & 76.52 & 69.34 & 19.62 & 14.60 & 12.79 \\
			\SetRow{bg=green!25}
			$\checkmark$ & $-$ & $\checkmark$ & $\checkmark$ &  $\checkmark$  & $\checkmark$  & \textbf{80.71} & 91.22 & 74.95 & 69.14 & \textbf{25.50} & \textbf{17.68} & \textbf{14.75} \\
			\hline[1.25pt]
		\end{tblr}
	}
	\end{minipage}
\end{table}%

%

We compare the effectiveness of various real-time optimization strategies, experimental results are described in Table~\ref{tab:opt}. For real-time settings (\textless 100ms), each ground truth frame has a corresponding predicted frame, making it an end-to-end approach. For non-real-time settings (\textgreater 100ms), we employ Kalman filter~\cite{kalman1960new, li2020towards, wang2023we} to forecast and interpolate unmatched intermediate frames. Hence, Table~\ref{tab:opt} can also be viewed as a comparison between end-to-end methods and traditional SOTA methods. The table reveals that all optimization strategies mentioned in Appendix~\ref{app:opt} effectively reduce model latency with minimal impact on offline accuracy, decreasing from the initial 263.33ms to 80.71ms. Concurrently, the sAP$\rm _{3D}$ gradually increases due to the reduced latency, rising from 2.53\% to 23.73\% in the easy level, achieving an improvement of over 20\%. Therefore, we choose the configuration highlighted in \textcolor{green}{green row} in Table~\ref{tab:opt} as the final optimization strategy.

\subsection{Comparison of the Domain Gap}
\label{app:gap}
\begin{wraptable}{r}{0.5\textwidth}
	\centering
	\vspace*{-1em}
	\caption{Comparison of the domain gap between our split Tracking dataset and the 3D Object Detection dataset.}
	\vspace*{-0.5em}
	\label{tab:app_gap}
	\resizebox*{\linewidth}{!}{
		\begin{tblr}{colspec=ccc||ccc, stretch=1.,  rows={ht=1.25\baselineskip}}
			\hline[1.25pt]
			\SetCell[r=2]{c} {Method} &\SetCell[r=2]{c} {Sensor} &\SetCell[r=2]{c} {Dataset} & \SetCell[c=3]{c} AP$\rm _{3D}$ & & \\
			\cline{2-7}
			&  &  &  Easy & Mod. & Hard \\
			\hline
			\hline
			PointPillars~\cite{lang2019pointpillars}  & LiDAR & Object Detection & 87.75	&78.38	&75.18  \\
			PointPillars~\cite{lang2019pointpillars}  & LiDAR & our split Tracking & 94.57	&88.35	&84.85  \\			
			DSGN++~\cite{chen2022dsgn++} & Stereo & Object Detection & 83.63	&66.41	&61.38  \\
			DSGN++~\cite{chen2022dsgn++} & Stereo & our split Tracking & 91.79	&78.35	&69.79  \\
			\hline[1.25pt]
		\end{tblr}%
	}

\end{wraptable}
We compare the domain gap between our split tracking dataset, which consists of 4,291 training samples and 3,672 validation samples, and the widely recognized KITTI 3D Object Detection dataset~\cite{geiger2012we}, comprising 3,712 training samples and 3,769 validation samples. Specifically, we train and test PointPillar~\cite{lang2019pointpillars} and DSGN++~\cite{chen2022dsgn++} on both datasets, and then compare the AP$\rm_{3D}$
for the Car category at \( IoU=0.7 \). The experimental results are shown in the Table~\ref{tab:app_gap}. From the table, we observe that both methods perform better on our split Tracking dataset, indicating a smaller domain gap compared to the Object Detection dataset. However, given that we have 579 additional training samples and that accuracy may further decrease under streaming perception constraints, this difference is deemed reasonable.

\begin{wraptable}{r}{0.5\textwidth}
	\centering
	\vspace*{-1em}
	\caption{Ablation studies of fusion stage.}
	\vspace*{-0.5em}
	\label{tab:fusion}
	\resizebox*{\linewidth}{!}{
		\begin{tblr}{colspec=c||ccc|ccc, stretch=1.,  rows={ht=1.25\baselineskip}}
			\hline[1.25pt]
			\SetCell[r=2]{c} {Fusion  Stage} & \SetCell[c=3]{c} sAP$\rm _{BEV}$ & & & \SetCell[c=3]{c} sAP$\rm _{3D}$ & & \\
			\cline{2-7}
			& Easy & Mod. & Hard & Easy & Mod. & Hard \\
			\hline
			\hline
			before head~\cite{yang2022real}  & 80.72 & 68.51 & 63.14 & 73.02 & 58.37 & 51.86 \\
			before BEV backbone & 83.20 & 71.15 & 65.74 & 75.38 & 59.55 & 53.11 \\
			\hline[1.25pt]
		\end{tblr}%
	}
	\vspace*{-1em}
	
\end{wraptable}

%
%
\subsection{Impact of Fusion Stage} 
\label{app:fusion_stage}

To validate the hypothesis about insufficient receptive field of shallow detection heads, we conducted comparative experiments by setting the fusion stage before the detection heads and before the BEV backbone, respectively. The comparison results are described in Table~\ref{tab:fusion}. The table demonstrates that fusing historical features before the BEV backbone, rather than preceding the detection head as proposed in StreamYOLO~\cite{yang2022real}, leads to competitive improvements. Specifically, the improvements at the three difficulty levels exceed 2\%, 1\%, and 1\%, respectively.

\begin{table}[htbp]
	\begin{minipage}[t]{0.52\textwidth}
		\centering
		\caption{Comparison with SOTA fusion method and ablation studies on FFF. The term ``$r_{d}$'' and ``$d$'' represent the downsample ratio and maximum displacement, respectively.}
		\label{tab:fff}
		\resizebox{1\linewidth}{!}{
			\begin{tblr}{colspec=cc|c|ccc|ccc, stretch=1.,  rows={ht=1.25\baselineskip}}
				\hline[1.25pt]
				 \SetCell[r=2]{c} {$r_{d}$} & \SetCell[r=2]{c} {$d$} & \SetCell[r=2]{c} {Latency \\ (ms)} & \SetCell[c=3]{c}{sAP$\rm _{BEV}$} & & & \SetCell[c=3]{c}{sAP$\rm _{3D}$} & & \\
				\cline{4-9}
				& & & Easy & Mod. & Hard & Easy & Mod. & Hard \\
				\hline
				\hline
				\SetCell[c=9]{c} \textit{Dual-Flow}~\cite{yang2022real} \\ 
				\hline
				 $-$ & $-$ & 2.67 & 83.20 & 71.15 & 65.74 & 75.38 & 59.55 & 53.11 \\
				\hline
				\SetCell[c=9]{c} \textit{Feature-Flow Fusion (Ours)} \\ 
				\hline
				\SetRow{bg=green!25}
				2 & 3 & 7.67 & 85.08 & 71.98 & 66.49 & 75.88 & 61.57 & 54.77 \\
				2 & 4 & 12.19 & 85.50 & 71.68 & 66.13 & 76.82 & 60.73 & 54.02 \\
				2 & 5 & 14.63 & 85.48 & 71.93 & 66.22 & 75.91 & 59.83 & 54.46 \\
				4 & 3 & 6.26 & 83.27 & 71.35 & 64.24 & 76.11 & 59.40 & 53.82 \\
				\hline[1.25pt]
			\end{tblr}
		}
	\end{minipage}
	\hfill
	\begin{minipage}[t]{0.41\textwidth}
		\centering
		\caption{Grid search of $\tau$ in Equation~\ref{eq:tau} for MCL.}
		\label{tab:mcl}
		\resizebox{1\linewidth}{!}{
			\begin{tblr}{colspec={c||ccc|ccc}, stretch=1., rows={ht=1.25\baselineskip},}
				\hline[1.25pt]
				\SetCell[r=2]{c} {$\tau$} & \SetCell[c=3]{c} sAP$\rm _{BEV}$ & & & \SetCell[c=3]{c} sAP$\rm _{3D}$ & & \\
				\cline{2-7}
				& Easy & Mod. & Hard & Easy & Mod. & Hard \\
				\hline
				\hline
				\SetCell[c=7]{c} \textit{w/o MCL} \\
				\hline
				$-$ & 83.20 & 71.15 & 65.74 & 75.38 & 59.55 & 53.11 \\
				\hline
				\SetCell[c=7]{c} \textit{w/ MCL} \\
				\hline
				0.0 & 84.99 & 71.71 & 66.04 & 76.13 & 61.26 & 54.40 \\  
				0.2 & 83.43 & 71.82 & 66.10 & 76.72 & 61.88 & 55.08 \\
				0.4 & 83.35 & 71.17 & 64.35 & 76.30 & 61.58 & 54.82 \\
				0.6 & 84.68 & 71.63 & 65.85 & 76.30 & 61.68 & 54.74 \\
				\SetRow{bg=green!25}
				0.8 & 84.63 & 71.01 & 66.18 & 78.15 & 62.28 & 56.80 \\
				1.0 & 85.09 & 71.49 & 66.03 & 75.74 & 59.62 & 54.24 \\
				\hline[1.25pt]
			\end{tblr}	
		}
	\end{minipage}
	\vspace{-0.5em}

%
\end{table}

\subsection{Latency and Hyperparameters of FFF}
\label{app:fff}
We report the latency and accuracy of FFF under different feature downsampling ratios and maximum search displacement in Table~\ref{tab:fff}. The Table shows that with a downsampling ratio of 2 and maximum displacement of 3, FFF achieves the best performance while only taking 7.67ms (highlighted in \textcolor{green}{green}). At this setting, sAP$\rm _{3D}$ shows a 2.01\% improvement at the moderate level compared to the SOTA Dual-Flow~\cite{yang2022real}.

\subsection{Value of \( \tau \)}
\label{app:mcl}
We get the value of hyperparameter \( \tau \) in Equation~\ref{eq:tau} through grid search. Experimental results are shown in Table~\ref{tab:mcl}. We observe that when only utilizing the velocity loss setting in MCL ($\tau=0$), the improvements at the three difficulty levels compared to the scheme without MCL are about 1\%, 1.5\%, and 1\%, respectively. Further, when introducing the acceleration loss, the performance reaches its maximum improvement when the balancing parameter $\tau=0.8$  (highlighted in \textcolor{green}{green}). At this setting, the improvements at the three difficulty levels are 2.77\%, 2.73\%, and 3.69\%, respectively.

\subsection{Complexity of LKBB}
\label{app:lkbb}

\begin{wraptable}{r}{0.6\textwidth}
	\centering
	\vspace{-1.5em}
	\caption{Comparison of the Complexity of BEV Backbones.}
	\vspace{-0.5em}
	\label{tab:lkbb}
	\resizebox{\linewidth}{!}{
		\begin{tblr}{colspec={c|ccc|ccc|ccc}, stretch=1., rows={ht=1.25\baselineskip},}
			\hline[1.25pt]
			\SetCell[r=2]{c}{Method}	& \SetCell[r=2]{c} {Params \\ (M)} & \SetCell[r=2]{c} {FLOPs \\ (G)} & \SetCell[r=2]{c}{Latency \\ (ms)} & \SetCell[c=3]{c} sAP$\rm _{BEV}$ & & & \SetCell[c=3]{c} sAP$\rm _{3D}$ & & \\
			\cline{5-10}
			& & & & Easy & Mod. & Hard & Easy & Mod. & Hard \\
			\hline \hline
			Hourglass (original) & 0.794 & 17.737  & 3.66 & 83.20 & 71.15 & 65.74 & 75.38 & 59.55 & 53.11 \\
			LKBB (ours) & 0.810 & 11.695 & 6.73 & 84.97 & 71.96 & 66.53 & 76.74 & 60.31 & 55.01 \\
			\hline[1.25pt]
		\end{tblr}
	}
\end{wraptable}
We present in Table~\ref{tab:lkbb} a comparison between LKBB and the original hourglass structure~\cite{newell2016stacked} in terms of parameter count, computational complexity, and latency. Our LKBB has a similar parameter count to the original structure, yet it reduces computational complexity by nearly 6 GFLOPs. At this juncture, the overhead incurred is merely an additional approximate 3ms, while achieving a competitively enhanced accuracy.

\end{document}